\documentclass[11pt,a4paper]{article}
\usepackage[hyperref]{emnlp2020}
\usepackage{times}
\usepackage{latexsym}
\usepackage{amsmath}
\usepackage{amssymb}
\usepackage{graphicx}
\usepackage{caption}
\usepackage{subcaption}
\usepackage{float}
\usepackage{multirow}
\usepackage{booktabs}
\usepackage{natbib}
\usepackage{todonotes}

\usepackage{microtype}

\aclfinalcopy 

\title{Tradeoffs in Sentence Selection Techniques \\for Open-Domain Question Answering}

\author{Shih-Ting Lin    {\normalfont ~and~} Greg Durrett \\
  The University of Texas at Austin \\ \texttt{\{j0717lin,gdurrett\}@cs.utexas.edu}
  \\}

\date{}

\begin{document}
\maketitle
\begin{abstract}
Current methods in open-domain question answering (QA) usually employ a pipeline of first retrieving relevant documents, then applying strong reading comprehension (RC) models to that retrieved text. However, modern RC models are complex and expensive to run, so techniques to prune the space of retrieved text are critical to allow this approach to scale. In this paper, we focus on approaches which apply an intermediate sentence selection step to address this issue, and investigate the best practices for this approach. We describe two groups of models for sentence selection: QA-based approaches, which run a full-fledged QA system to identify answer candidates, and retrieval-based models, which find parts of each passage specifically related to each question. We examine trade-offs between processing speed and task performance in these two approaches, and demonstrate an ensemble module that represents a hybrid of the two. From experiments on Open-SQuAD and TriviaQA, we show that very lightweight QA models can do well at this task, but retrieval-based models are faster still. An ensemble module we describe balances between the two and generalizes well cross-domain.
\end{abstract}

\section{Introduction}
\label{sec:intro}
\begin{figure}[t]
\vspace{0.0cm}
\centering
\includegraphics[width=78mm ]{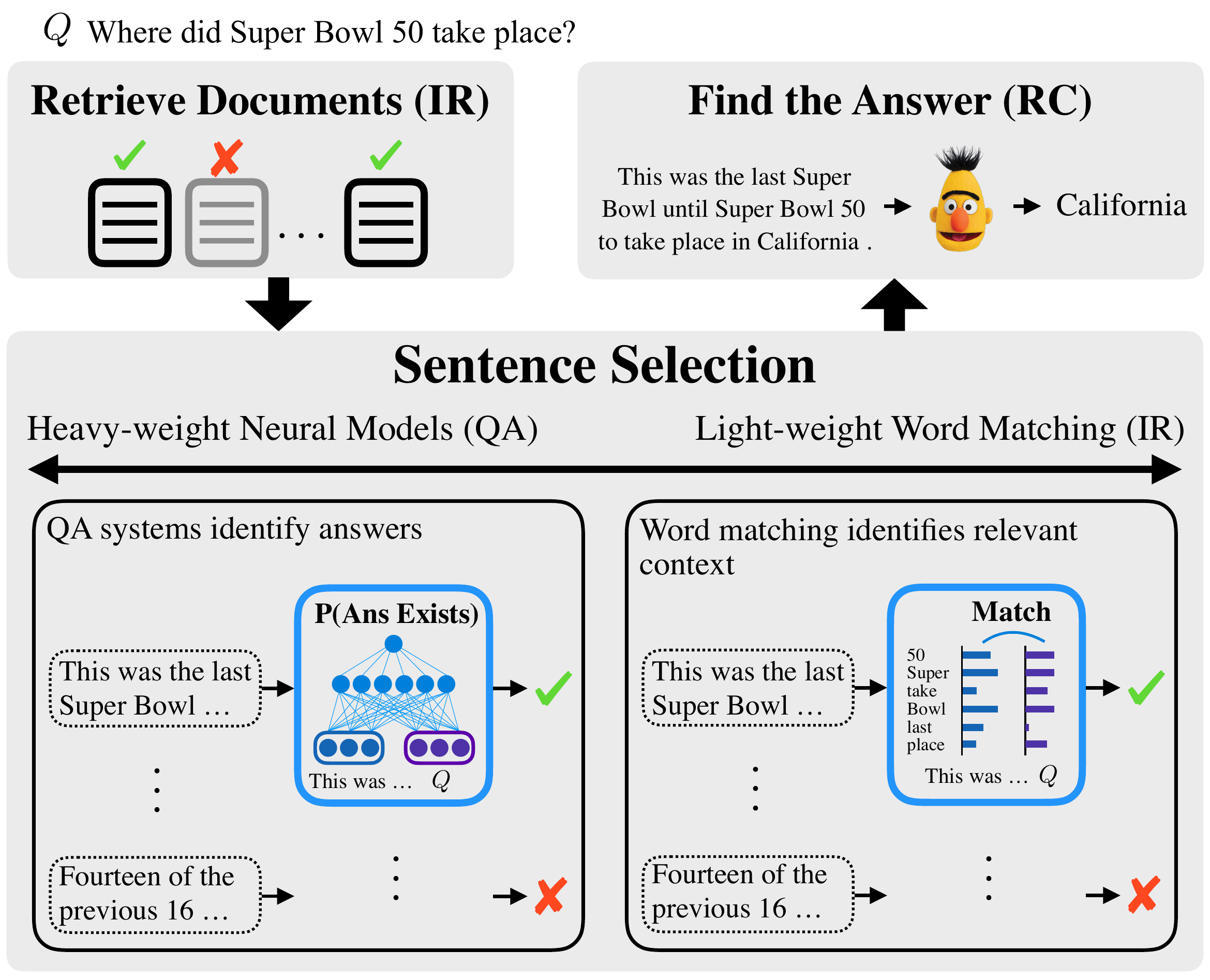}
\caption{An overview of our coarse-to-fine open-domain QA pipeline with a sentence selection step.}
\label{fig:overview}
\vspace{-0.5cm}
\end{figure}

Open-domain question answering (QA) systems commonly apply a two-stage process: first use a light-weight information retrieval (IR) module to obtain documents related to the question, then apply a reading comprehension (RC) model to get the answer \cite{drqa, R-3, das2018multistep, tfidf-para}. This IR-RC pipeline lets a system easily scale to a database with millions of documents. The RC model used in this pipeline is typically a high-capacity neural model \cite{bidaf, match-lstm, bert}, but running a computationally-expensive RC model even in this pipeline setting becomes expensive when retrieving dozens of documents \cite{bertserini, multi-passage-bert}. 
Dense indexing of documents can be one solution to this issue, but this either impairs the model's ability to effectively recognize candidate answer contexts \cite{DSIndex, Karpukhin2020DensePR}, or requires a complicated training scheme \cite{ORQA}. A more attractive approach, which balances simplicity, speed, and accuracy, is a coarse-to-fine pipeline (Figure~\ref{fig:overview}): use a stronger model to select sentences before the RC model to further narrow down its search space \cite{deep-cascade, Wu2019IntegratedTF, raiman-miller-2017-globally}.

In this paper, we investigate the best practices for the sentence selection method, which should effectively identify sentences relevant to the answer without introducing much overhead \cite{MudrakartaTSD18, fastqa}. We start by identifying two groups of existing approaches based on the model complexity, as in Figure~\ref{fig:overview}: first, QA-based modules that model this process with complex QA-like architectures \cite{minimal, min-etal-2017-question}, and second, light-weight retrieval-based modules that view this process as a retrieval task \cite{coarse-to-fine}. 
In addition, we present an ensemble selector that hybridizes these two types of methods in an efficient way so as to achieve a better balance between the accuracy and the speed. Finally these modules are evaluated in terms of accuracy, latency, and how well they generalize to a new open QA setting.

We conduct experiments on two open-domain QA settings: Open-SQuAD \cite{squad,drqa} and TriviaQA \cite{joshi-etal-2017-triviaqa}. Our results demonstrate that QA-based selectors achieve the best performance in terms of downstream accuracy, but retrieval-based selectors are much faster, albeit with a cost to accuracy. Surprisingly, stronger QA systems give little benefit over minimal, efficient ones. Our ensemble model strikes a balance between speed and accuracy, and from cross-dataset generalization experiments, we see that the ensemble model is exploiting a more domain-agnostic strategy to identify plausible answer contexts in the IR-RC pipeline.

\section{Techniques for Sentence Selection}
We now describe the sentence selection modules we study in our coarse-to-fine pipeline. Formally, the input to our selection module is a question $\mathbf{q}$ and a set of documents $\{\mathbf{d}^1, ..., \mathbf{d}^K\}$ retrieved by the IR system, each with sentences $\{\mathbf{x}^{k1},..., \mathbf{x}^{kn}\}$. Note that $\mathbf{q}$, $\mathbf{d}^k$, and $\mathbf{x}^{ki}$ are sequences of tokens. Our task is to assign each sentence a score $S(\mathbf{q},\mathbf{x}^{ki})$; based on these scores, we will select the final subset of sentences to feed into the RC model.

We consider techniques for modeling $S$ to identify sentences likely to contain the answer. Note that here we only consider the ``black-box'' sentence selection techniques, as opposed to those relying on particular implementations of the RC model \cite{raiman-miller-2017-globally, Wu2019IntegratedTF}.

\subsection{QA-based Selection Modules}

One class of approaches to sentence selection tries to directly identify those likely to contain the answer with QA models which are simpler and faster than the final RC component of the pipeline. A model may be too simple to do downstream QA well but could still identify a reasonable sentence set. Here, we explore three different systems. 

\paragraph{BiDAF}
BiDAF \cite{bidaf} assigns an answer probability to each span in the textual context via a product of two distributions, one modeling answer starts and the other modeling answer ends. To adapt this model for sentence selection, we first feed the document $\mathbf{d}^k$ into the model along with $\mathbf{q}$. We then take the logit of the best token span within $\mathbf{x}^{ki}$ as its selection score. Intuitively, this model scores a sentence based on how good the most probable answer candidate in that sentence looks. 
To train the BiDAF selector, we use the standard closed-domain QA datasets, then directly apply it in the open-domain setting.


While BiDAF is significantly smaller than BERT-based QA models \cite{bert}, its reliance on RNNs still makes it expensive to run. We also investigate a \textbf{BiDAF-small} selector with the hidden size reduced by 90\% so as to speed up inference. 

\paragraph{FastFusionNet}
FastFusionNet \cite{fastfusionnet} is a QA model that exploits parallelism to achieve fast inference while maintaining high accuracy. Because this model effectively balances speed and performance, we also adopt it as one of our underlying architectures for QA-based selector. We can use it to compute selection scores in the same way as BiDAF since its output layers are the same.

\subsection{Retrieval-based Selection Modules}

We consider two retrieval-based methods. We compute the cosine similarity between the \textbf{tf-idf} vector of the question and a sentence, and use it as the selection score. Second, we consider using \textbf{bag-of-words} embeddings to encode semantic information of both the question and a targeted sentence; the resulting dense representations are then passed to a feed-forward network to get the selection score.

\subsection{The Ensemble Selection Module}
The two groups of modules described above represent two major factors in sentence selection: while QA-based modules exploit complex networks to select a high-quality sentence set, the retrieval-based methods aim to achieve a fast selection at the cost of reducing model capacity. Since these two characteristics are both keys to a high-performance open-domain QA system, here we develop an approach that combines QA-based and retrieval-based selectors to take both of their advantages.

First, we simplify the QA-based selector into an \textbf{\textsc{AnsFind}} module to more directly check whether any phrase within the sentence is a plausible answer to the question. Our architecture for the \textsc{AnsFind} module is similar to \citet{fastqa}, although we are using the model specifically for classification of plausible answer spans. We form a set of candidate answers $\{c_i\}$ on each sentence $\mathbf{x}$ by extracting all of its constituents from a constituency parser. Then, we compute the representation $\mathbf{h}_i$ for each $c_i$ by simple vector averaging and feedforward operations. For the question vector $\mathbf{h_q}$, we apply a RNN only over the prefix of the question up to the first named entity (the wh-phrase, plus potentially verbs which may be indicative of answer type), since we are just trying to determine whether this span is a \emph{plausible} candidate answer. The probability of $c_i$ being a plausible answer, $P(c_i|\mathbf{q}, \mathbf{x}^{ki})$, is then computed by passing $\mathbf{h_q}$ and $\mathbf{h}_i$ to a feed-forward network. Finally, the module uses a binary selection score $S_{\text{ans}}(\mathbf{q}, \mathbf{x}^{ki})$, which is 1 if there exists any $c_i$ such that $P(c_i|\mathbf{q}, \mathbf{x}^{ki})$ is larger than a threshold, and 0 otherwise. Like QA-based selectors, the training of this module is also done with closed-domain QA datasets. See the Appendix for the detailed architecture and training procedure.

Second, we implement an \textbf{\textsc{EvdMatch}} module, borrowing ideas from retrieval-based selectors to determine whether evidence mentioned in the question can be found in a sentence $\mathbf{x}^{ki}$. First, we use a constituency parser to collect base constituents $\{u_j\}$ from the question (one level up from the leaves). Then we compute a binary score indicating if there exists some span of tokens in $\mathbf{x}^{ki}$ that matches $u_j$. The final selection score for a sentence $S_{\text{evd}}(\mathbf{q}, \mathbf{x}^{ki})$ is the summation of the matching score for all evidence. This computation is analogous to tf-idf and so is very fast, while also capturing an effective QA strategy that may generalize cross-domain. See the Appendix for details.

The \textsc{AnsFind} and \textsc{EvdMatch} modules serve as the two components of our ensemble selector. To combine the results from them, we compute the final selection score $S(\mathbf{q}, \mathbf{x}^{ki}) = S_{\text{ans}}(\mathbf{q}, \mathbf{x}^{ki}) + S_{\text{evd}}(\mathbf{q}, \mathbf{x}^{ki})$ for each sentence $\mathbf{x}^{ki}$. Even though these pieces rely on a parser, compare to the QA-based selectors, the ensemble module is less computationally intensive since it only relied on cheap operations such as feed-forward layers and simple string matching.

\section{Coarse-to-fine Pipeline}
\label{sec:impl}
We use Wikipedia as our corpus for the open QA setting. We segment each page into paragraphs and treat these paragraphs as ``documents.'' In our pipeline, we first utilize Anserini\footnote{\url{http://anserini.io/}} to retrieve 50 documents given the question; note that improved retrieval methods like \citet{Karpukhin2020DensePR} and \citet{DSIndex} could be exploited here as well. Then we apply our sentence selectors to pick 10 sentences with highest selection scores. The sentences in the subset are concatenated and passed to a BERT-based QA model, which is fine-tuned on closed-domain QA datasets, to get the final answer. To measure the latency, we run our pipelines on Amazon EC2 p3.2xlarge instances, which are equipped with a NVIDIA Tesla V100 GPU.

\section{Experimental Results}

We evaluate on two English open-domain QA datasets. (1) \textbf{Open-SQuAD} \cite{squad}: following previous work \cite{bertserini}, we treat the questions in SQuAD dev set as queries and ignore the provided contexts to form a open-domain QA dataset. (2) \textbf{TriviaQA} \cite{triviaqa}: this dataset is designed for open-domain QA. We take the questions in the unfiltered set for experiments.

\begin{figure}[t]
\centering
\includegraphics[width=78mm]{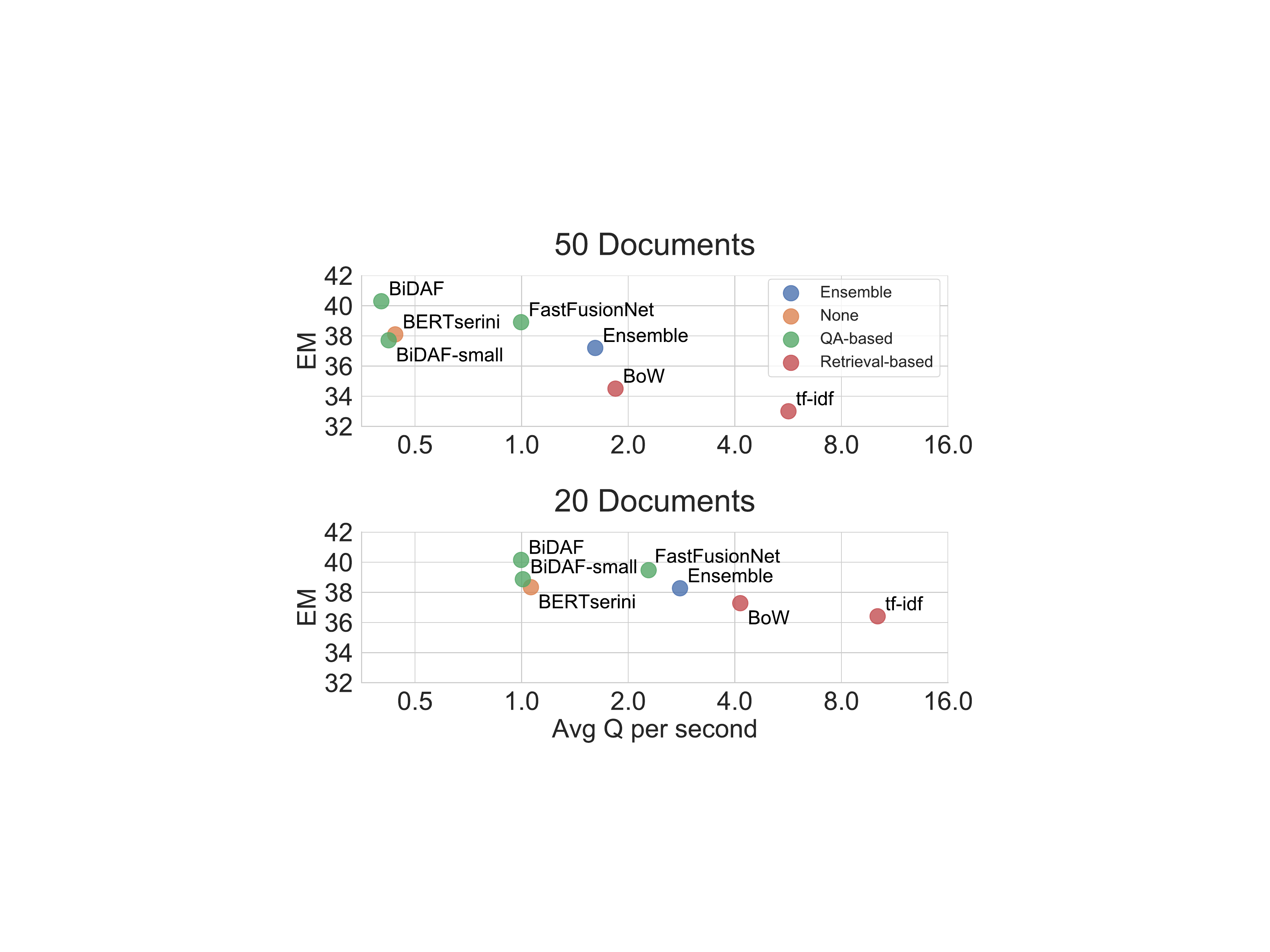}
\caption{The downstream exact match (EM) score and the averaged number of processed question per second on TriviaQA for our sentence selectors along with BERTserini, which is a standard IR-RC approach, when input with 50 and 20 documents respectively. The averaged processing time is measured with 100 randomly sampled questions from the dataset.}
\label{fig:tqa_speed}
\vspace{-0.2cm}
\end{figure}

\subsection{Trade-off between Speed and Accuracy}
We first evaluate the sentence selectors we study in terms of speed and accuracy, which are two key criteria in open-domain QA.  The results are illustrated in Figure~\ref{fig:tqa_speed}; graphs for SQuAD are in the appendix. BERTserini \cite{bertserini}, which is a standard IR-RC system without sentence selection, is also implemented here as a baseline. All three QA-based selectors achieve a comparable task performance with BERTserini and outperform other types of selectors, which indicates these modules are able to select a high-quality sentence set for the following RC model. However, the BiDAF selector fails to provide any speed-up in the coarse-to-fine pipeline. The BiDAF-small selector is also as slow as BERTserini; when using GPUs, reducing model capacity is not a good solution for speeding up the selection with a fixed architecture. On the other hand, the FastFusionNet successfully achieves more than 2x speed-up over BERTserini.

Second, we examine the results of retrieval-based selectors. This class of selectors is significantly faster: our bag-of-words module is about 2 times faster than FastFusionNet and the tf-idf module achieves over 4x speed-up. However, this efficiency is at cost of an accuracy drop (about a 4-point decrease comparing to FastFusionNet on 50 documents). 
Finally, our ensemble selector demonstrates a balance between the speed and accuracy: it achieves a comparable task performance with the QA-based selectors and even makes the pipeline faster than the one with FastFusionNet.

In sum, the QA-based modules achieve the best task performance, whereas the retrieval-based modules can do selection more efficiently, with the ensemble method and FastFusionNet balancing these.

\begin{table}[t]
\small
\centering
\renewcommand{\tabcolsep}{0.9mm}
\begin{tabular}{ l | c  c  c | c  c  c}
\toprule
Evaluation Set & \multicolumn{3}{c|}{SQ} & \multicolumn{3}{c}{TQA} \\
\midrule
Selector Training Set & SQ & TQA & $\Delta$ & TQA & SQ & $\Delta$ \\ 
\toprule
BiDAF Selector & 42.2 & 29.6 & -12.6 & 40.3 & 39.9 & -0.4 \\ 
FastFusionNet Selector & 42.2 & 30.4 & -11.8 & 38.9 & 41.1 & \textbf{2.17} \\ 
\midrule
BoW-Selector & 30.6 & 27.7 & -2.9 & 34.5 & 34.5 &  \phantom{0}0.0 \\
tf-idf-Selector & 30.1 & 30.1 & \phantom{0}0.0 & 33.0 & 33.0 & \phantom{0}0.0 \\
\midrule
Ensemble & 41.5 & 40.9 & \textbf{-0.6} & 37.2 & 37.1 & -0.1 \\
\bottomrule
\end{tabular}
\caption{Comparison of generalization ability between various selection modules on Open-SQuAD (SQ) and TriviaQA (TQA) dataset. The metric shown here is the final QA Exact Match on 50 documents.}
\vspace{-0.0cm}
\label{tab:gen}
\end{table}

\subsection{Generalization Ability}
Next, we investigate the ability of these selectors to generalize across domains. In this experiment, the RC model is trained in-domain but the sentence selectors are trained out-of-domain; this allows us to assess how general each technique is. The results are shown in Table \ref{tab:gen}. First, we can see from the right column that both of the QA-based selectors are able to generalize well from SQuAD to the TriviaQA dataset, with FastFusionNet even doing better than when trained on TriviaQA. However, when transferring from TriviaQA to SQuAD, a huge performance drop can be seen, likely due to the noise in the TriviaQA training data. In contrast, the ensemble selector as well as the retrieval-based selectors barely suffer from the domain shift between SQuAD and TriviaQA dataset, suggesting that retrieval-based methods (including the \textsc{EvdMatch} module) can generalize well and that the \textsc{AnsFind} module learns a more domain-agnostic strategy than standard QA models. 

\begin{table}[t]
\small
\centering
\renewcommand{\tabcolsep}{1.0mm}
\begin{tabular}{ l | c  c  c | c  c  c}
\toprule
 & \multicolumn{3}{c|}{SQ} & \multicolumn{3}{c}{TQA} \\
\midrule
   & Recl & EM & F1 & Recl & EM & F1 \\ 
\toprule
DrQA & -- & 29.8 & -- & -- & 37.5 & 42.2 \\
R\textsuperscript{3} & -- & 29.1 & 37.5 & -- & -- & --\\
DENSPI-Hybrid & -- & 36.2 & 44.4 & -- & -- & --\\
ORQA & -- & 26.5 & -- & -- & 45.1 & -- \\
Multi-Step Reader & -- & -- & -- & -- & 39.8 & 44.3 \\
\toprule
\multicolumn{7}{c}{50 documents}\\
\midrule
BERTserini (ours) & 83.2 & 38.3 & 44.7 & 80.3 & 38.1 & 43.5 \\
\midrule
\multicolumn{7}{c}{50 documents $\rightarrow$ 10 sentences} \\
\midrule
BiDAF Selector & 67.5 & \textbf{42.2} & 49.7 & 68.9 & \textbf{40.3} & 45.8 \\ 
BiDAF-small Selector & 65.8 & 42.0 & 49.5 & 64.5 & 37.7 & 42.8 \\ 
FastFusionNet Selector & 67.8 & 42.2 & 50.0 & 67.1 & 38.9 & 44.4 \\ 
\midrule
BoW-Selector & 48.9 & 30.6 & 37.5 & 58.2 & 34.5 & 40.1 \\
tf-idf-Selector & 46.3 & 30.1 & 36.7 & 49.9 & 33.0 & 38.6  \\
\midrule
Ensemble & 62.7 & \textbf{41.5} & 48.8 & 61.0 & 37.2 & 42.3 \\
 Only \textsc{EvdMatch} & 60.9 & 40.7 & 48.1 & 58.7 & 36.5 & 41.5 \\
 Only \textsc{AnsFind} & 54.6 & 36.0 & 42.9 & 60.8 & \textbf{38.3} & 43.8 \\
\bottomrule
\end{tabular}
\caption{Results on Open-SQuAD (SQ) and TriviaQA (TQA) dataset. ``Recl" stands for the recall of gold answer appearing in the input to the final RC model. EM and F1 are the final open-domain QA result.}
\vspace{-0.5cm}
\label{tab:squad_results}
\end{table}

\subsection{Comparison to Previous Work}

In Table \ref{tab:squad_results}, we compare our coarse-to-fine pipelines to existing open-domain QA systems. The results indicate that our pipelines with QA-based and ensemble selectors achieve comparable performance with most of the existing systems listed here except ORQA \cite{ORQA}, which achieves stronger performance on TriviaQA. Note that ORQA uses a BERT-based retrieval module, which can potentially be exploited in our pipelines as mentioned in Section \ref{sec:impl}. Finally, our ablations shows that the \textsc{AnsFind} module alone (Only \textsc{AnsFind}) is better than the retrieval-based selectors on SQuAD and achieves an even better performance than both the ensemble selector and BERTserini on TriviaQA. While the questions in TriviaQA are somewhat different from those in SQuAD, the answer finding strategy is still effective for selection.




\bibliographystyle{acl_natbib}
\bibliography{emnlp2020}

\newpage
~
\newpage
\appendix



\section{Supplementary Material}
\paragraph{Detailed Architecture of the \textsc{AnsFind} Module}
As mentioned in the main text, we first form a set of candidate answers $\{c_i = \mathbf{x}_{b:e}^{ki}\}$, where $b, e$ are the start and end position of some constituent, for a sentence $\mathbf{x}^{ki}$ using an off-the-shelf constituency parser; note that this step can be done offline in advance. Then, we compute the span representation for each candidate $c_i$. Concretely, sentence $\mathbf{x}^{ki}$ is first passed to pretrained GloVe \cite{glove} and a CNN-based character embedding layer to get the word embedding matrix $\mathbf{E} \in \mathbb{R}^{L\times d}$, where $d$ is the word embedding dimension. The span representation for $c_i$, is then extracted by averaging over the rows of $\mathbf{E}_{b:e}$. We also augment the span representation with the context words around the candidate so that the model can get more clues about its semantics. In practice, we include four words on each side of the candidate span to form the final span representation $\mathbf{h}_i$. Then we feed the wh-phrase of the question $\mathbf{q}$ into a RNN and extract the final state as the the question vector $\mathbf{h_q} \in \mathbb{R}^d$. The probability of $c_i$ being a plausible answer is then computed by:
\begin{align}
    \mathbf{h}_{qc} &= [\mathbf{h_q}; \mathbf{h}_i; \mathbf{h}_i\odot \mathbf{h_q}; \mathbf{E}_b; \mathbf{E}_e] \\
    \mathbf{h}'_{qc} &= \mathrm{FFNN}(\mathbf{h}_{qc}) \\
    P(c_i|\mathbf{q}, \mathbf{x}^{ki}) &= \mathrm{sigmoid}(\mathbf{v}^{\top} \mathbf{h}'_{qc})
\end{align}
where $\mathrm{FFNN}$ is a two layer feed-forward network.

\paragraph{Detailed Architecture of the \textsc{EvdMatch} Module}
First, we collect ``evidence'' from the question by splitting it into multiple parts. We perform this splitting by running the same off-the-shelf constituency parser used in \textsc{AnsFind} module on the question, and treat the base constituents (constituents which are directly parents of POS tags) as the ``evidence set'' $\{u_j\}$. In this manner, we can get a syntactically-informed set of concepts from the question that are relatively fine-grained. Then, for each evidence $u_j$, we first lemmatize and lowercase it, then output a binary score indicating if there exists some span of tokens in $\mathbf{x}$ that matches $u_j$. To deal with the fact that entity evidence may be mentioned by a pronoun coreferent with a previous sentence, in practice we expand the current sentence to include its previous one to do the matching.  Finally, the selection score for a sentence $\mathbf{x}^{ki}$ is computed as:
\begin{align}
     m(u_j, \mathbf{x}^{ki}) &= 
    \begin{cases}
        1, u_j \in \mathbf{x}^{ki} \\
        0, \; \mathrm{otherwise}
    \end{cases} \\
    S_{evd}(\mathbf{x}^{ki}) &= \sum_{u_j}m(u_j, \mathbf{x}^{ki}) + m(u_j, \mathbf{x}^{k\{i-1\}})
\end{align}

\paragraph{Training Details}
For our QA-based selectors, we first train them with standard closed-domain QA datasets with standard hyperparameters, then apply them in the open-domain setting. For the \textsc{AnsFind} module in our ensemble selector, we also first train it in a closed QA setting. To get the training instances for binary classification, we first find the constituent with highest Jaccard similarity with the gold answer, and treat it as a positive example. We randomly sample 20 other constituents as negatives. To combat the data imbalance, we upweight the loss on positive instances by a factor of 5 during training. We set the classification threshold to 0.3 during test time to retrieve a high-recall set of answer candidates.

\begin{figure}[t]
\centering
\includegraphics[width=78mm]{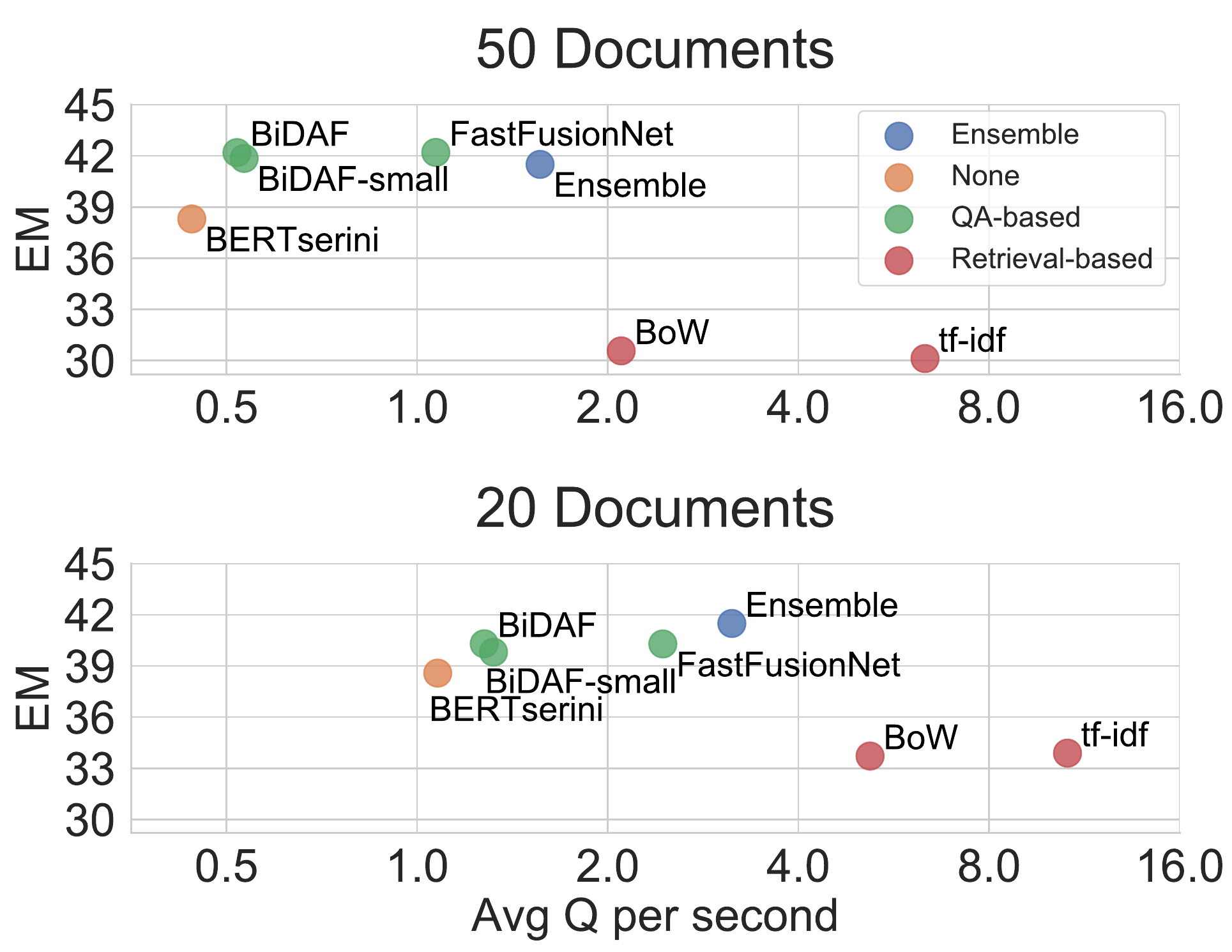}
\caption{The downstream exact match (EM) score and the averaged number of processed question per second on SQuAD for our sentence selectors along with BERTserini when input with 50 and 20 documents respectively. The averaged processing time is also measured with 100 randomly sampled questions from the dataset. Retrieval-based methods are faster, QA-based methods are more accurate, and the ensemble hybridizes these.}
\label{fig:squad_speed}
\vspace{-0.2cm}
\end{figure}




\end{document}